\documentclass{article} % For LaTeX2e
\usepackage{nips15submit_e,times}
\usepackage{hyperref}
\usepackage{url}

\usepackage{amsmath,graphicx,booktabs}
\usepackage{epstopdf,amssymb}
\usepackage{multirow}
\title{VoxResNet:  Deep Voxelwise Residual Networks for Volumetric Brain Segmentation} %Autocontex
\author{
Hao Chen, Qi Dou, Lequan Yu and Pheng-Ann Heng %\thanks{ Use footnote for providing further information}
\\
Department of Computer Science and Engineering\\
The Chinese University of Hong Kong\\
Hong Kong, China \\
\texttt{hchen@cse.cuhk.edu.hk} \\
}

\nipsfinalcopy % Uncomment for camera-ready version

\begin{document}

\maketitle

\begin{abstract}
Recently deep residual learning with residual units for training very deep neural networks advanced the state-of-the-art performance on 2D image recognition tasks, e.g., object detection and segmentation.
However, how to fully leverage contextual representations for recognition tasks from volumetric data has not been well studied, especially in the field of medical image computing, where a majority of image modalities are in volumetric format.
In this paper we explore the deep residual learning on the task of volumetric brain segmentation.
There are at least two main contributions in our work.
First, we propose a deep voxelwise residual network, referred as~\emph{VoxResNet}, which borrows the spirit of deep residual learning in 2D image recognition tasks, and is extended into a 3D variant for handling volumetric data.
%To tackle it efficiently and effectively,
Second, an auto-context version of VoxResNet is proposed by seamlessly integrating the low-level image appearance features, implicit shape information and high-level context together for further improving the volumetric segmentation performance.
Extensive experiments on the challenging benchmark of brain segmentation from magnetic resonance (MR) images corroborated the efficacy of our proposed method in dealing with volumetric data.
%demonstrated that VoxResNet enjoys the benefit of increasing depths from network, which
We believe this work unravels the potential of 3D deep learning to advance the recognition performance on volumetric image segmentation.
\end{abstract}

\section{Introduction}
Over the last few years, deep learning especially deep convolutional neural networks (CNNs) have emerged as one of the most prominent approaches for image recognition problems in various domains including computer vision~\cite{krizhevsky2012imagenet,simonyan2014very,long2015fully,szegedy2015going} and medical image computing~\cite{prasoon2013deep,ronneberger2015u,chen2015standard,shin2016deep}.
Most of the studies focused on the 2D object detection and segmentation tasks, which have shown a compelling accuracy compared to previous methods employing hand-crafted features.
However, in the field of medical image computing, volumetric data accounts for a large portion of medical image modalities, such as Computed Tomography (CT) and Magnetic Resonance Imaging (MRI), etc.
Furthermore, the volumetric diagnosis from temporal series usually requires data analysis even in a higher dimension.
In clinical practice, the task of volumetric image segmentation plays a significant role in computer aided diagnosis (CADx), which provides quantitative measurements and aids surgical treatments.
Nevertheless, this task is quite challenging due to the high-dimensionality and complexity along with volumetric data.
To the best of our knowledge, there are two types of CNNs developed for handling volumetric data.
The first type employed modified variants of 2D CNN by taking aggregated adjacent slices~\cite{chen2015automatic} or orthogonal planes (i.e., axial, coronal and sagittal)~\cite{prasoon2013deep,roth2014new} as input to make up complementary spatial information. However, these methods cannot make full use of the contextual information sufficiently, hence it is not able to segment objects from volumetric data accurately.
Recently, other methods based on 3D CNN have been developed to detect or segment objects from volumetric data and demonstrated compelling performance~\cite{dou2016automatic,kamnitsas2016efficient,li2014deep}. %ji20133d
Nevertheless, these methods may suffer from limited representation capability using a relatively shallow depth or may cause optimization degradation problem by simply increasing the depth of network.
Recently, deep residual learning with substantially enlarged depth further advanced the state-of-the-art performance on 2D image recognition tasks~\cite{he2015deep,he2016identity}.
Instead of simply stacking layers, it alleviated the optimization degradation issue by approximating the objective with residual functions.

Brain segmentation for quantifying the brain structure volumes can be of significant value on diagnosis, progression assessment and treatment in a wide range of neurologic diseases such as Alzheimer's disease~\cite{giorgio2013clinical}. Therefore, lots of automatic methods have been developed to achieve accurate segmentation performance in the literature.
Broadly speaking, they can be categorized into three classes:
1) Machine learning methods with hand-crafted features. These methods usually employed different classifiers with various hand-crafted features, such as support vector machine (SVM) with spatial and intensity features~\cite{van2013automated,moeskops2015evaluation}, random forest with 3D Haar like features~\cite{wang2015links} or appearance and spatial features~\cite{pereira2016automatic}. These methods suffer from limited representation capability for accurate recognition.
2) Deep learning methods with automatically learned features. These methods learn the features in a data-driven way, such as 3D convolutional neural network~\cite{cciccek20163d}, parallelized long short-term memory (LSTM)~\cite{stollenga2015parallel}, and 2D fully convolutional networks~\cite{nie2016fully}.
These methods can achieve more accurate results while eliminating the need for designing sophisticated input features. Nevertheless, more elegant architectures are required to further advance the performance.
3) Multi-atlas registration based methods~\cite{aljabar2009multi,artaechevarria2009combination,sarikaya2013multi}. However, these methods are usually computationally expensive, hence limits its capability in applications requiring fast speed.

To overcome aforementioned challenges and further unleash the capability of deep neural networks, we propose a deep voxelwise residual network, referred as~\emph{VoxResNet}, which borrows the spirit of deep residual learning to tackle the task of object segmentation from volumetric data.
Extensive experiments on the challenging benchmark of brain segmentation from volumetric MR images demonstrated that our method can achieve superior performance, outperforming other state-of-the-art methods by a significant margin.

\section{Method}
\subsection{Deep Residual Learning}
Deep residual networks with residual units have shown compelling accuracy and nice convergence behaviors on several large-scale image recognition tasks, such as ImageNet~\cite{he2015deep,he2016identity} and MS COCO~\cite{dai2015instance} competitions.
By using identity mappings as the skip connections and after-addition activation, residual units can allow signals to be directly propagated from one block to other blocks. Generally, the residual unit can be expressed as following
%\begin{gather}\label{residual}
%  y_l = x_l + \mathcal{F}(x_l, W_l)
%%  h(x_l) = x_l \\
%%  x_{l+1} = y_l
%\end{gather}
%By setting $h$ as an identity mapping: $h(x_l) = x_l$, we can get
\begin{gather}\label{residual2}
  x_{l+1} = x_l + \mathcal{F}(x_l, W_l)
\end{gather}
here the $\mathcal{F}$ denotes the residual function, $x_l$ is the input feature to the $l$-th residual unit and $W_l$ is a set of weights correspondingly associated with the residual unit.
The key idea of deep residual learning is to learn the additive residual function $\mathcal{F}$ with respect to the input feature $x_l$.
Hence by unfolding above equation recursively, the $x_L (L>l\ge1)$ can be derived as
\begin{gather}\label{residual2}
  x_L = x_l + \sum_{i=l}^{L-1} \mathcal{F}(x_i, W_i)
\end{gather}
therefore, the feature $x_L$ of any deeper layers can be represented as the feature $x_l$ of shallow unit $l$ plus summarized residual functions $ \sum_{i=l}^{L-1} \mathcal{F}(x_i, W_i)$.
The derivations imply that residual units can make information propagate through the network smoothly.

\begin{figure}
\centering
  \includegraphics[width=1.\linewidth]{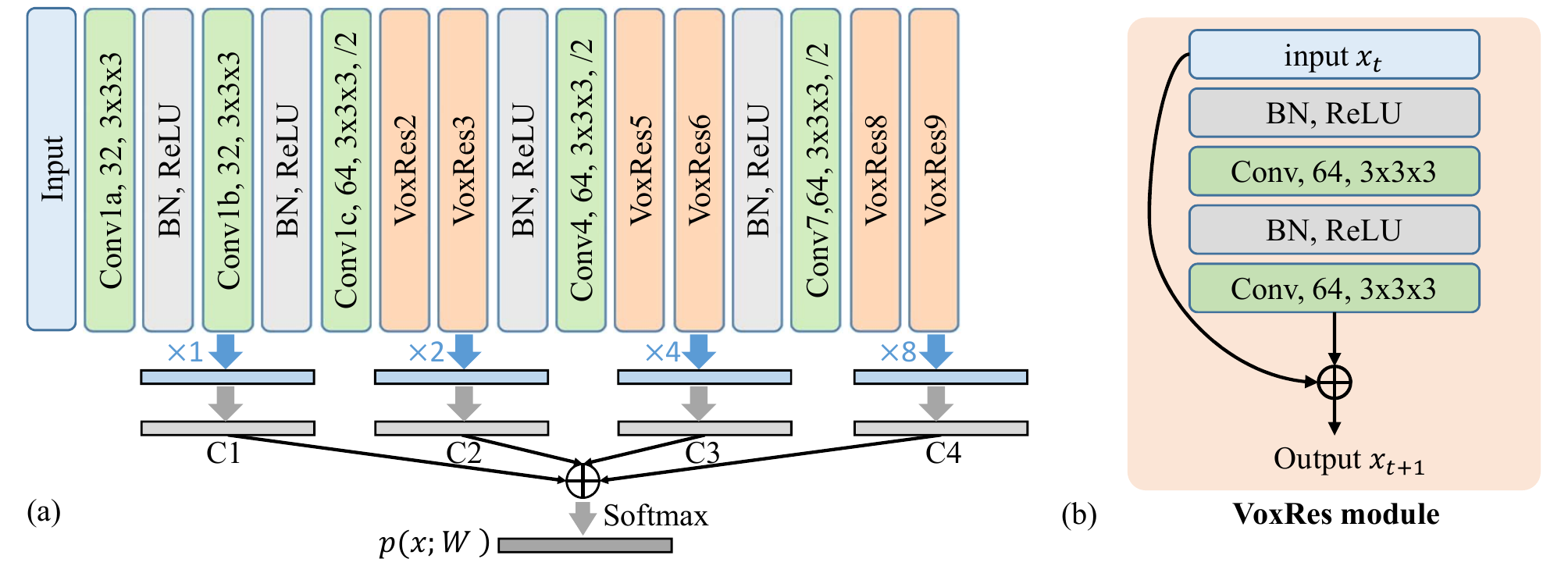}
\caption{(a) The architecture of proposed VoxResNet for volumetric image segmentation; (b) The illustration of VoxRes module.}
\label{fig:network}
\end{figure}

\subsection{VoxResNet for Volumetric Image Segmentation}
Although 2D deep residual networks have been extensively studied in the domain of computer vision~\cite{he2015deep,he2016identity,shen2016weighted,zagoruyko2016wide}, to the best of our knowledge, seldom studies have been explored in the field of medical image computing, where the majority of dataset is volumetric images.
In order to leverage the powerful capability of deep residual learning and tackle the object segmentation tasks from high-dimensional volumetric images efficiently and effectively, we extend the 2D deep residual networks into a 3D variant and design the architecture following the spirit from~\cite{he2016identity} with full pre-activation, i.e., using asymmetric after-addition activation, as shown in Figure~\ref{fig:network}(b).

The architecture of our proposed VoxResNet for volumetric image segmentation is shown in Figure~\ref{fig:network}(a).
Specifically, it consists of stacked residual modules (i.e., VoxRes module) with a total of 25 volumetric convolutional layers and 4 deconvolutional layers~\cite{long2015fully}, which is the deepest 3D convolutional architecture so far.
In each VoxRes module, the input feature $x_l$ and transformed feature $\mathcal{F}(x_l, W_l)$ are added together with skip connection as shown in Figure~\ref{fig:network}(b), hence the information can be directly propagated in the forward and backward passes.
Note that all the operations are implemented in a 3D way to strengthen the volumetric feature representation learning.
Following the principle from VGG network~\cite{simonyan2014very} and deep residual networks~\cite{he2016identity}, we employ small convolutional kernels (i.e., $3\times3\times3$) in the convolutional layers, which have demonstrated the state-of-the-art performance on image recognition tasks.
%Anisotropic kernel for differen resolutions in volumetric data.
Three convolutional layers are along with a stride 2, which reduced the resolution size of input volume by a factor of 8. This enables the network to have a large receptive field size, hence enclose more contextual information for improving the discrimination capability.
Batch normalization layers are inserted into the architecture intermediately for reducing internal covariate shift~\cite{ioffe2015batch}, hence accelerate the training process and improve the performance.
In our network, the rectified linear units, i.e., $f(x) = max(0,x)$, are utilized as the activation function for non-linear transform~\cite{krizhevsky2012imagenet}.

There is a huge variation of 3D anatomical structure shape, which demands different suitable receptive field sizes for better recognition peformance.
In order to handle the large variation of shape sizes, multi-level contextual information (i.e., 4 auxiliary classifiers C1-C4 in Figure~\ref{fig:network}(a)) with deep supervision~\cite{lee2015deeply,chen2016deep} is fused in our framework. Therefore, the whole network is trained in an end-to-end way by minimizing following objective function with standard back-propagation
\begin{gather}
  \mathcal{L}(x, y;\theta) = {\lambda} \psi(\theta)-\sum_{\alpha} \sum_{x\in \mathcal{V}} \sum_{c} w_{\alpha} y_c^x \log p_c^\alpha(x;\theta) - \sum_{x\in \mathcal{V}} \sum_{c} y_c^x \log p_c(x;\theta) \label{loss}
\end{gather}
where the first part is the regularization term ($L_2$ norm in our experiments) and latter one is the fidelity term consisting of auxiliary classifiers and final target classifier.
The tradeoff of these terms is controlled by the hyperparameter $\lambda$. % which is set as 0.01 heuristically in our experiments.
$w_{\alpha}$ (where $\alpha$ indicates the index of auxiliary classifiers) is the weights of auxiliary classifiers, which were set as 1 initially and decreased till marginal values (i.e., $10^{-3}$) in our experiments.
The weights of network are denoted as $\theta=\{W\}$, $p_c(x; \theta)$ or $p_c^\alpha(x; \theta)$ denotes the predicted probability of $c$th class after softmax classification layer for voxel $x$ in volume space $\mathcal{V}$, and $y_c^x \in\{0,1\}$ is the corresponding ground truth. 
, i.e., $y_c^x=1$ if voxel $x$ belongs to the $c$th class, otherwise 0.
\begin{figure}
\centering
  \includegraphics[width=.85\linewidth]{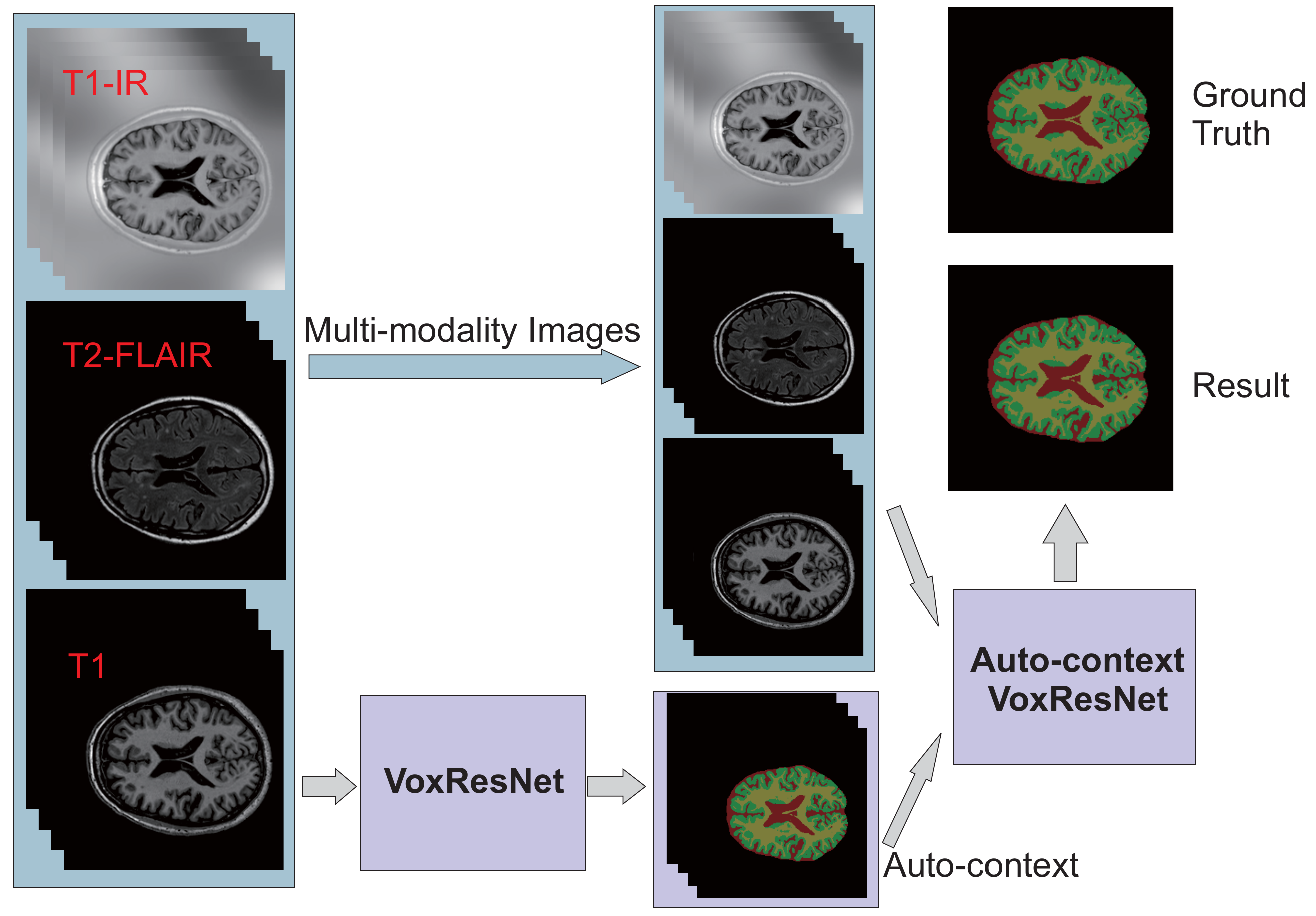}
\caption{An overview of our proposed framework for fusing auto-context with multi-modality information.}
\label{fig:framework}
\end{figure}

%\subsection{3D conditional random field}
\subsection{Multi-modality and Auto-context Information Fusion} %
In medical image computing, the volumetric data are usually acquired with multiple imaging modalities for robustly examining different tissue structures.
For example, three imaging modalities including T1, T1-weighted inversion recovery (T1-IR) and T2-FLAIR are available in brain structure segmentation task~\cite{mendrik2015mrbrains} and four imaging modalities are used in brain tumor (T1, T1 contrast-enhanced, T2, and T2-FLAIR MRI)~\cite{menze2015multimodal} and lesion studies (T1-weighted, T2-weighted, DWI and FLAIR MRI)~\cite{maier2017isles}.
The main reason for acquiring multi-modality images is that the information from multi-modality dataset can be complementary, which provides robust diagnosis results.
Thus, we concatenate these multi-modality data as input, then the complementary information is jointly fused during the training of network in an implicit way, which demonstrated consistent improvement compared to any single modality.
%The overview of whole framework is illustrated in Figure~\ref{}.
%Region-based consistency.

Furthermore, in order to harness the integration of high-level context information, implicit shape information and original low-level image appearance for improving recognition performance, we formulate the learning process as an auto-context algorithm~\cite{tu2008auto}.
Compared with the recognition tasks in computer vision, the role of auto-context information can be more important in the medical domain as the anatomical structures are roughly positioned and constrained~\cite{tu2010auto}.
Different from~\cite{tu2010auto} utilizing the probabilistic boosting tree as the classifier, we employ the powerful deep neural networks as the classifier.
Specifically, given the training images, we first train a~\emph{VoxResNet} classifier on original training sub-volumes. Then the discriminative probability maps generated from~\emph{VoxResNet} are then used as the context information, together with the original volumes (i.e., appearance information), to train a new classifier~\emph{Auto-context VoxResNet}, which further refines the semantic segmentation and removes the outliers.
Different from the original auto-context algorithm in an iterative way~\cite{tu2010auto}, our empirical study showed that following iterative refinements gave marginal improvements. Therefore, we chosen the output of~\emph{Auto-context VoxResNet} as the final results.

\section{Experiments}
%\subsection{MRBrainS challenge}
\subsection{Dataset and Pre-processing}
We validated our method on the MICCAI MRBrainS challenge data, an ongoing benchmark for evaluating algorithms on the brain segmentation.
The task of MRBrainS challenge is to segment the brain into four-class structures, i.e., background, cerebrospinal fluid (CSF), gray matter (GM) and white matter (WM).
The datasets were acquired at the UMC Utrecht of patients with diabetes and matched controls with varying degrees of atrophy and white matter lesions~\cite{mendrik2015mrbrains}. %(with increased cardiovascular risk)
Multi-sequence 3T MRI brain scans, including T1-weighted, T1-IR and T2-FLAIR, are provided for each subject.
The training dataset consists of five subjects with manual segmentations provided.
%Manual segmentations (ground truth) were drawn on the thick-slice scans (3mm slice thickness), using an in-house developed manual segmentation tool based on the contour segmentation objects (CSO) tool available in Mevislab.
The test data includes 15 subjects with ground truth held out by the organizers for independent evaluation.

In the pre-processing step, we subtracted Gaussian smoothed image and applied Contrast-Limited Adaptive Histogram Equalization (CLAHE) for enhancing local contrast by following~\cite{stollenga2015parallel}.
Then six input volumes including original images and pre-processed ones were used as input data in our experiments.
We normalized the intensities of each slice with zero mean and unit variance before inputting into the network.

%\begin{figure}
%\centering
%  \includegraphics[width=.8\linewidth]{figures/stats}
%\caption{Statistics of lesion size.}
%\label{fig:stats}
%\end{figure}
\begin{figure}
\centering
  \includegraphics[width=1.0\linewidth]{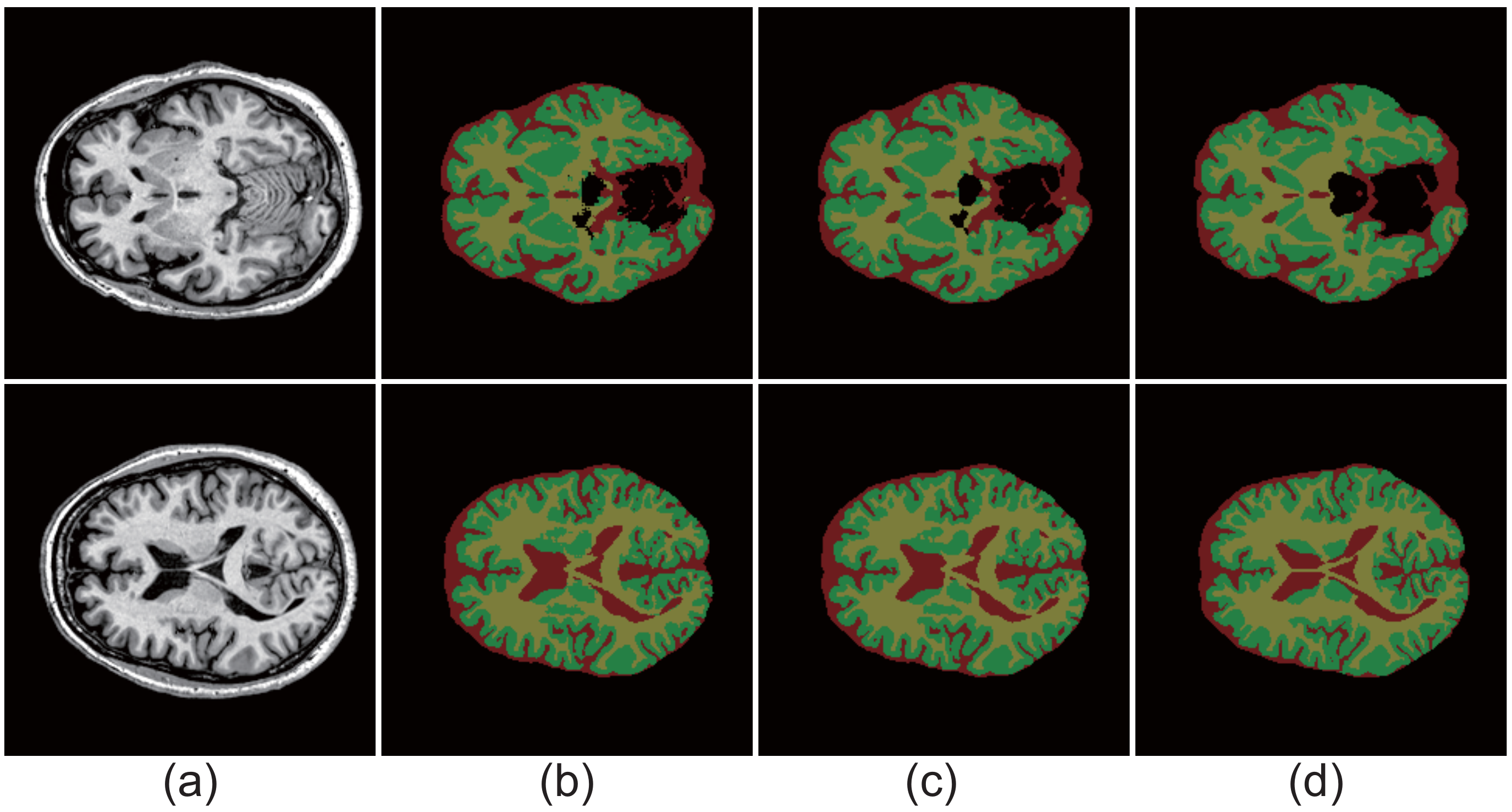}
\caption{The example results of validation data (yellow, green, and red colors represent the WM, GM, and CSF, respectively): (a) original MR images, (b) results of VoxResNet, (c) results of Auto-context VoxResNet, (d) ground truth labels.}
\label{fig:results}
\end{figure}
\subsection{Evaluation and Comparison}
The evaluation metrics of MRBrainS challenge include the Dice coefficient (DC), the 95th-percentile of the Hausdorff distance (HD) and the absolute volume difference (AVD), which are calculated for each tissue type (i.e., GM, WM and CSF), respectively~\cite{mendrik2015mrbrains}.
The details of evaluation can be found in the challenge website\footnote{MICCAI MRBrainS Challenge: \url{http://mrbrains13.isi.uu.nl/details.php}}.
%The final ranking is based on the evaluation results of all 15 test datasets and is determined as follows: For each evaluation measure (DC, HD-95, AVD), the mean value over all 15 datasets is determined for WM, GM and CSF.
%Each team receives a rank (1=best) for each tissue type (GM, WM, CSF) and each evaluation measure (DC, HD-95, AVD) based on the mean value of the evaluation measures over all 15 datasets.
%The final score is determined by adding the ranks of all tissue types and evaluation measures for each team.
%The team with the lowest score will be ranked number 1. In case two teams would have an equal score, the team with the lowest standard deviation over the tissue types will be ranked number 1.

To investigate the efficacy of multi-modality and auto-context information, we performed extensive ablation studies on the validation data (leave one out cross-validation).
The detailed results of cross-validation are reported in Table~\ref{table:cross-result}.
We can see that combining the multi-modality information can dramatically improve the segmentation performance than that of single image modality, especially on the metric of DC, which demonstrates the complementary characteristic of different imaging modalities.
Moreover, by integrating the auto-context information, the performance of DC can be further improved.
The qualitative results of brain segmentation can be seen in Figure~\ref{fig:results} and we can see that the results of combining multi-modality and auto-context information can give more accurate results visually than only multi-modality informaltion.

\begin{table}[th]
\centering
\caption{Cross-validation Results of MR Brain Segmentation using Different Modalities (DC: \%, HD: mm, AVD: \%).}
\resizebox{1.0\linewidth}{!}{
\begin{tabular}{ | l | l | l | l | l | l | l | l | l | l | }
\hline
 \multirow{2}{*}{Modality} & \multicolumn{3}{c|}{GM} & \multicolumn{3}{c|}{WM} & \multicolumn{3}{c|}{CSF}  \\
 \cline{2-10}
	 & DC & HD & AVD & DC & HD & AVD & DC & HD & AVD  \\ \hline
	T1 & 86.96 & \bf{1.36}  & \bf{4.67}  &  89.70 & 1.92  & 6.85  & 79.58 & 2.71 & 17.55  \\ \hline
	T1-IR & 80.61  & 1.92  & 8.45  & 85.89  & 2.87  & 7.42  & 76.44  & 3.00  & 12.87  \\ \hline
%	T2-FLAIR & 81.13  & 1.92  &  9.15 & 83.01  & 3.00  & 1.53 & 75.34  &  3.00 &  1.90  \\ \hline
	T2-FLAIR & 81.13  & 1.92  &  9.15 & 83.21  & 3.00  & 4.99 & 75.34  &  3.03 &  \bf{3.77} \\ \hline
%	All & 87.46 & 0.958 & 4.61 & 89.87 & 1.36 & 2.88 & 81.87 & 2.14 & 6.51  \\ \hline
%	All(w/o res) & 87.46 & 0.96 & 5.36  & 90.21  & 1.36 & 2.43  & 82.88  & 1.92  & 5.11   \\ \hline
	All & 86.86 & \bf{1.36} & 7.13 & 90.22 &\bf{1.36} & 5.12 & 81.97 & \bf{2.14} & 9.87  \\ \hline
	All+auto-context & \bf{87.83} & \bf{1.36} & 6.22  & \bf{90.63} & \bf{1.36} & \bf{2.22} & \bf{82.76} & \bf{2.14}  & 5.50   \\ \hline
%	average(VoxResNet and auto-context) & 87.53 & 1.36 & 6.52  & 90.56 & 1.36 & 3.78 & 82.56 & 1.92  & 7.58   \\ \hline
%	All-res & 87.46 & 0.958 & 4.61 & 89.87 & 1.36 & 2.88 & 81.87 & 2.14 & 6.51  \\ \hline
\end{tabular}
}
%DC = Dice Coefficient, HD = 95th percentile of Hausdorff Distance, AVD = Absolute Volume Difference
\label{table:cross-result}
\end{table}

Regarding the evaluation of testing data, we compared our method with several state-of-the-art methods, including MDGRU, 3D U-net~\cite{cciccek20163d} and PyraMiD-LSTM~\cite{stollenga2015parallel}.
The MDGRU applied a neural network with the main components being multi-dimensional gated recurrent units and achieved quite good performance.
The 3D U-net extended previous 2D version~\cite{ronneberger2015u} into a 3D variant and highlighted the necessity for volumetric feature representation when applied to 3D recognition tasks.
The PyraMiD-LSTM parallelised the multi-dimensional recurrent neural networks in a pyramidal fashion and achieved compelling performance. The detailed results of testing data from different methods on brain segmentation from MR images can be seen in Table~\ref{table:result_final}.
We can see that deep learning based methods can achieve much better performance than hand-crafted feature based methods.
The results of VoxResNet (see CU\_DL in Table~\ref{table:result_final}) by fusing multi-modality information achieved better performance than other deep learning based methods, which demonstrated the efficacy of our proposed framework.
Incorporating the auto-context information (see CU\_DL2 in Table~\ref{table:result_final}) , the performance of DC can be further improved.
Overall, our methods achieved the top places in the challenge leader board out of 37 competing teams.

%\begin{table}
%\begin{center}
%\caption{The Dice Coefficient (DC) results of different methods in MRBrainS Challenge.}
%\label{table:result_dice}
%\begin{tabular}{ l l l }
%\toprule
%	Method & GM & WM \\ \midrule
%	CU\_DL(ours) & 86.12 & 0.7860 \\
%	CU\_DL2(ours) & 86.15 & 0.7810 \\
%	FBI/LMB Freiburg & 85.05 & 0.7856 \\
%	PyraMiD-LSTM2 & 84.89 & 0.7542 \\
%	MIAC &  85.05 & \bf{0.8001} \\
%	IDSIA & 84.82 & 0.7832 \\
%	ISI-Neonatology & 85.77 & 0.7647 \\
%	STH &  84.77 & 0.7152 \\
%	UNC-IDEA & 84.65  & 0.6166 \\
%	MNAB2 & 84.50 & 0.6543 \\
%\bottomrule
%\end{tabular}
%\end{center}
%\end{table}
%\begin{table}[th]
%\centering
%\caption{Cross-validation Results of MR Brain Segmentation using Different Modalities.}
%\begin{tabular}{ | l | l | l | l | }
%\hline
%	Modality & GM & WM & CSF \\ \hline
%	T1 & 0.838& 0.879 & 0.752 \\ \hline
%	T1-IR & \  & \  & \  \\ \hline
%	T2-FLAIR & \  & \  & \  \\ \hline
%	All & 0.875 & 0.899 & 0.819 \\ \hline
%\end{tabular}
%\label{table:multi-result}
%\end{table}

\begin{table}[th]
\centering
\caption{Results of MICCAI MRBrainS Challenge of different methods (DC: \%, HD: mm, AVD: \%. only top 10 teams are shown here).}
\resizebox{1.0\linewidth}{!}{
\begin{tabular}{ | l | l | l | l | l | l | l | l | l | l | c | }
\hline
 \multirow{2}{*}{Method} & \multicolumn{3}{c|}{GM} & \multicolumn{3}{c|}{WM} & \multicolumn{3}{c|}{CSF} & \multirow{2}{*}{Score*}  \\
 \cline{2-10}
	 & DC & HD & AVD & DC & HD & AVD & DC & HD & AVD &  \\ \hline
	CU\_DL (ours) & \bf{86.12} & \bf{1.47} & 6.42 & \bf{89.39} & \bf{1.94} & \bf{5.84} & 83.96 & 2.28 & 7.44 & \bf{39} \\ \hline
	CU\_DL2 (ours) & \bf{86.15} & \bf{1.45} & 6.60 & \bf{89.46} & \bf{1.94} & 6.05 & \bf{84.25} & 2.19 & 7.69 & \bf{39} \\ \hline
	MDGRU & 85.40 & 1.55 & 6.09 & 88.98 & 2.02 & 7.69 & 84.13 & 2.17 & 7.44 & 57 \\ \hline
	PyraMiD-LSTM2 & 84.89 & 1.67 & 6.35 & 88.53 & 2.07 & 5.93 & 83.05 & 2.30 & 7.17 & 59 \\ \hline
	FBI/LMB Freiburg~\cite{cciccek20163d} & 85.44 & 1.58 & 6.60 & 88.86 & 1.95 & 6.47 & 83.47 & 2.22 & 8.63 & 61 \\ \hline
	IDSIA~\cite{stollenga2015parallel} & 84.82 & 1.70 & 6.77 & 88.33 & 2.08 & 7.05 & 83.72 & \bf{2.14} & 7.09 & 77 \\ \hline
	STH & 84.77 & 1.71 & \bf{6.02} & 88.45 & 2.34 & 7.67 & 82.77 & 2.31 & \bf{6.73} & 86 \\ \hline
	ISI-Neonatology~\cite{moeskops2015evaluation} & 85.77 & 1.62 & 6.62 & 88.66 & 2.07 & 6.96 & 81.08 & 2.65 & 9.77 & 87 \\ \hline
	UNC-IDEA~\cite{wang2015links} & 84.36 & 1.62 & 7.04 & 88.68 & 2.06 & 6.46 & 82.81 & 2.35 & 10.5 & 90 \\ \hline
	MNAB2~\cite{pereira2016automatic} & 84.50 & 1.70 & 7.10 & 88.04 & 2.12 & 7.74 & 82.30 & 2.27 & 8.73 & 109 \\ \hline
\end{tabular}
}
%DC = Dice Coefficient, HD = 95th percentile of Hausdorff Distance, AVD = Absolute Volume Difference
\small{*Score = Rank DC + Rank HD + Rank AVD; a smaller score means better performance.}
\label{table:result_final}
\end{table}

\subsection{Implementation Details}
Our method was implemented using Matlab and C++ based on Caffe library~\cite{jia2014caffe,tran2015deep}.
It took about one day to train the network while less than 2 minutes for processing each test volume (size $240\times240\times48$) using a standard workstation with one NVIDIA TITAN X GPU.
Due to the limited capacity of GPU memory, we cropped volumetric regions (size $80\times80\times80\times m$, $m$ is number of image modalities and set as 6 in our experiments) for the input into the network. This was implemented in an on-the-fly way during the training, which randomly sampled the training samples from the whole input volumes.
In the test phase, the probability map of whole volume was generated in an overlap-tiling strategy for stitching the sub-volume results\footnote{Project page: \url{http://www.cse.cuhk.edu.hk/~hchen/research/seg_brain.html}}.

\section{Conclusions}

In this paper, we analyzed the capabilities of VoxResNet in the field of medical image computing and demonstrated its potential to advance the performance of biomedical volumetric image segmentation problems.
%Our method is inherently general and can be readily applied to other detection and segmentation problems.
The proposed method extends the deep residual learning in a 3D variant for handling volumetric data.
Furthermore, an auto-context version of VoxResNet is proposed to further boost the performance under an integration of low-level appearance information,  implicit shape information and high-level context.
Extensive experiments on the challenging segmentation benchmark corroborated the efficacy of our method when applied to volumetric data.
Moreover, the proposed algorithm goes beyond the application of brain segmentation and it can be applied in other volumetric image segmentation problems.
In the future, we will investigate the performance of our method on more object detection and segmentation  tasks from volumetric data.

\bibliographystyle{ieee}
\bibliography{reference}

\end{document}